\documentclass[letterpaper, 10 pt, conference]{ieeeconf}

\IEEEoverridecommandlockouts
\usepackage{amsmath,amsfonts}
\usepackage{algorithmic}
\usepackage{array}
\usepackage[caption=false,font=normalsize,labelfont=sf,textfont=sf]{subfig}
\usepackage{textcomp}
\usepackage{stfloats}
\usepackage{url}
\usepackage{verbatim}
\usepackage{graphicx}
\def\BibTeX{{\rm B\kern-.05em{\sc i\kern-.025em b}\kern-.08em
    T\kern-.1667em\lower.7ex\hbox{E}\kern-.125emX}}
\usepackage{balance}

\usepackage{comment}
\usepackage{xcolor}
\usepackage{multirow}
\usepackage{times}
\usepackage{epsfig}
\usepackage{amsmath}
\usepackage{amssymb}
\usepackage{bm}
\usepackage{siunitx}
\usepackage{graphics} 
\usepackage{graphicx}
\usepackage{float}
\usepackage{mathrsfs}
\usepackage{booktabs}
\usepackage{cite}

\definecolor{lightviolet}{rgb}{0.722, 0.549, 0.867}
\usepackage[pagebackref=false,breaklinks=true,colorlinks=true,bookmarks=false,linkcolor=orange,citecolor=cyan,urlcolor=lightviolet]{hyperref}

\newif\ifshowrevisions
\showrevisionsfalse
\definecolor{dayonecolor}{RGB}{0,105,95}
\definecolor{daytwocolor}{RGB}{25,75,150}
\definecolor{daythreecolor}{RGB}{0,110,65}
\definecolor{dayfourcolor}{RGB}{180,85,0}
\definecolor{dayfivecolor}{RGB}{120,45,140}
\definecolor{notationcolor}{RGB}{190,0,90}

\newcommand{\revthree}[1]{\ifshowrevisions\textcolor{daythreecolor}{#1}\else#1\fi}
\newcommand{\revfour}[1]{\ifshowrevisions\textcolor{dayfourcolor}{#1}\else#1\fi}

\newcommand{\revcolor}[1]{\ifshowrevisions\color{#1}\fi}
\newcommand{\notation}[1]{\ifshowrevisions\textcolor{notationcolor}{\bm{#1}}\else#1\fi}

\begin{document}


\title{\LARGE \bf
DiFF: Doppler-informed Flow Matching for Human Motion Flow
}

\author{
Kai Wang and Mingle Zhao
}

\maketitle



\begin{abstract}
Perceiving human motion via privacy-preserving 4D millimeter-wave (mmWave) radar is critical for next-generation human-robot interaction (HRI), where point cloud scene flow serves as a foundational motion representation. Yet the extreme sparsity and noise of 4D radar point clouds make non-rigid motion flow estimation severely ill-posed--a challenge that existing rigid-centric methods and prior works fail to adequately address, largely because they neglect the rich Doppler velocity cues inherent in 4D radar. We propose DiFF, a generative framework that marries Doppler-informed motion priors with a Kolmogorov-Arnold Network (KAN)-based conditional flow matching model. At its core, a KAN-attention mechanism enables expressive feature extraction, while a prior-guided generative process harnesses Doppler cues to regularize the ill-posed solution space. Extensive experiments show that DiFF achieves state-of-the-art (SOTA) performance across diverse real-world datasets, reducing 3D endpoint error to the millimeter scale on the mmBody benchmark. The source code is released at: \href{https://github.com/keroseus/DiFF/}{https://github.com/keroseus/DiFF/}.
\end{abstract}


\section{Introduction} \label{sec:intro}

Perceiving and understanding human motion is a critical capability for next-generation HRI, enabling safer human-robot collaboration, personalized assistive care, and intuitive smart home environments. Traditional approaches often rely on cameras or wearable sensors. However, these technologies are not only susceptible to environmental factors like poor illumination, but more importantly, they raise significant privacy concerns, rendering them unsuitable for deployment in private spaces. In light of these challenges, researchers have begun exploring alternative sensing modalities \cite{adib2013see}.

Single-chip mmWave radar has emerged as a low-cost and highly integrated sensor. Due to its Multiple-Input-Multiple-Output (MIMO) transceiver design, mmWave radar can provide reliable point clouds of a scene and effectively adapt to environmental dynamics. Building on these advantages, single-chip radars have seen widespread adoption and have been extensively researched in human sensing, spanning applications from vital sign monitoring and motion recognition \cite{Lien2016soli} to fine-grained gesture recognition \cite{jin2024Gesture}.

    \begin{figure}[!t]
        \centering
        \includegraphics[width=0.47\textwidth]{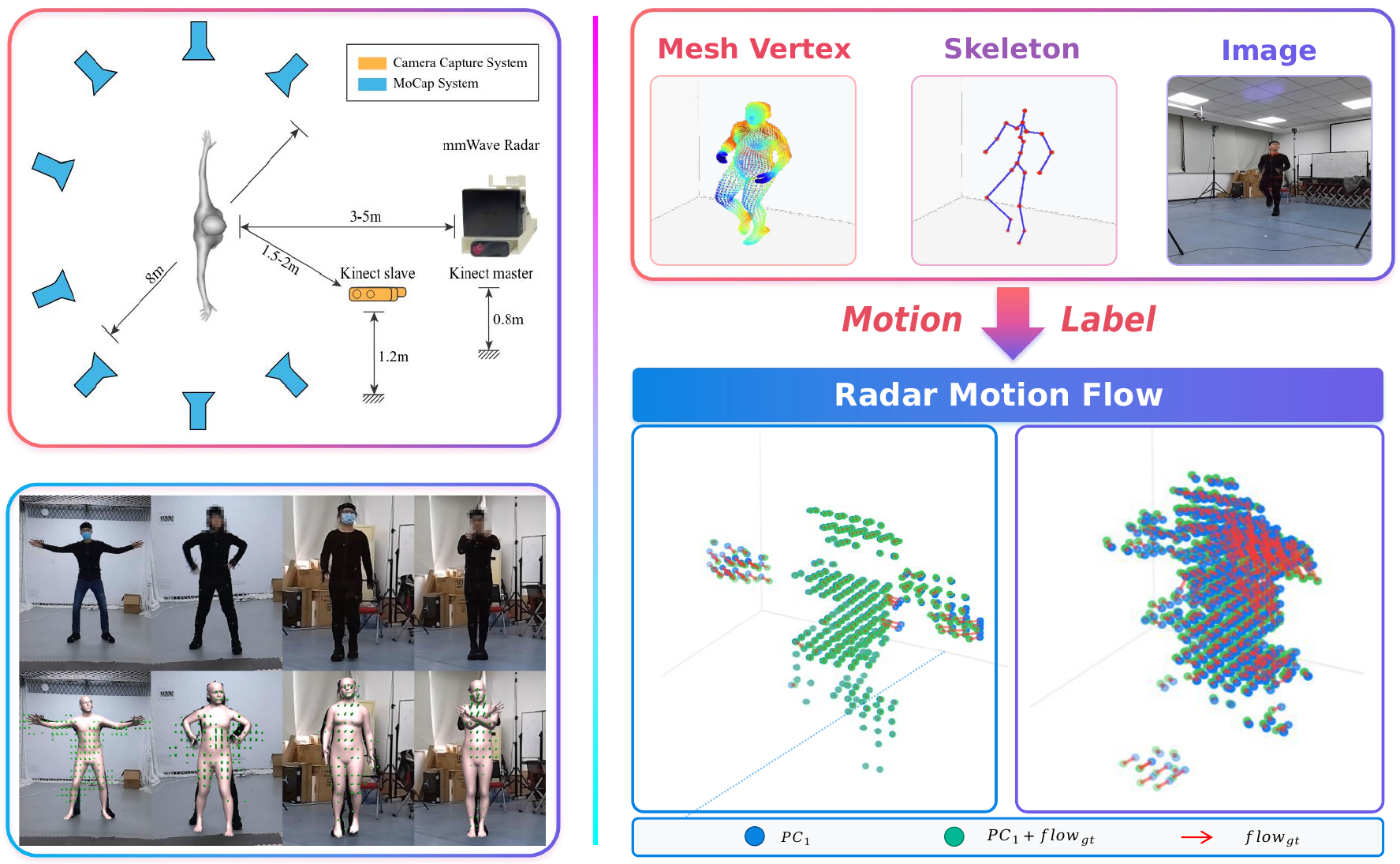}
        \caption{Data pipeline overview. (Left) Dataset setup and multi-modal annotations from \cite{chen2022mmbody}. (Right) Motion label generation from mesh vertices, skeleton, and images, and the proposed Doppler-informed radar motion flow.}
        \label{fig:physical:physical_vision}
    \end{figure}

In the context of human motion perception, milliFlow \cite{ding2024_milliflow} reveals that accurate scene flow estimation is crucial for a robot to better comprehend dynamic human behaviors. Nevertheless, extracting fine-grained motion information from sparse and noisy point clouds from low-cost radars is a formidable challenge. The inherent data sparsity makes establishing reliable point-to-point correspondences between consecutive scans exceptionally difficult, a problem exacerbated by the non-rigid and complex nature of human articulation. This intrinsic data ambiguity renders the estimation of human motion flow from sparse radar point clouds a severely ill-posed problem.

However, a limitation of previous works is that they could not take advantage of Doppler velocity information, as this feature is not available on the Vayyar mmWave radar in \cite{ding2024_milliflow}. A growing body of research shows that Doppler information inherent in 4D Radar/LiDAR sensors provides crucial additional constraints for key tasks such as motion sensing, perception, and estimation \cite{harlow2024new, DICP, gu2022learning, wu2022picking, yoon2023need, zhao2024fmcw, zhao2024free, khoche2026dogflow}, thus helping to address several long-standing and highly challenging problems at the sensor level. To fill this gap, we introduce DiFF, a novel framework that integrates Doppler-informed motion prior with Kolmogorov-Arnold Network (KAN) conditioned flow matching for estimating high-fidelity human motion flow. Our core innovation lies in combining an efficient flow matching paradigm with a powerful conditioning feature extractor network, KAN-Attention. This network is built upon the principle of KAN and incorporates a global attention structure. By replacing fixed activation functions with learnable rational polynomial functions, our conditioning network can adaptively capture complex geometric and temporal correlations within sparse point clouds, guiding the flow matching process to regress an accurate velocity field. The experimental results demonstrate that DiFF is a fast and accurate framework for estimating human motion flow through 4D mmWave radars. The contributions are:
    \begin{enumerate}
        \item To the best of our knowledge, this work is the first to introduce a flow matching network for human motion flow estimation on point clouds. Our method achieves SOTA accuracy and overall performance, reducing the motion flow prediction error to the millimeter level.
    
        \item We propose a KAN-based point cloud feature extractor. Combined with a global attention mechanism, the proposed architecture can capture both local geometric structures and global contextual information more effectively, while also improving inference efficiency.
    
        \item We design a Doppler-informed motion prior module, leveraging the Doppler information embedded in the 4D radar to estimate an initial radial flow, which serves as a physics-plausible guidance for flow matching. This design facilitates faster convergence and higher estimation accuracy. Building upon this, we incorporate a dynamic conditioning mechanism, which further enhances training stability.
    \end{enumerate}


\section{Related Work} \label{sec:related_work} 
\subsection{Scene Flow in Rigid and Deformable Scenes}
Scene flow estimation on point clouds has been predominantly driven by advancements in autonomous driving, where LiDAR sensors provide high-resolution data. Early pioneering works like FlowNet3D \cite{liu2019_flownet3d} and PointPWC-Net \cite{wu2020_pointpwcnet} established end-to-end deep learning frameworks by constructing cost volumes to learn point correspondences. Subsequent research has significantly improved performance by designing more sophisticated correlation mechanisms and network architectures. For instance, PV-RAFT \cite{wei2021_pvraft} introduced point-voxel correlation fields to enhance long-range motion modeling. To reduce the reliance on large-scale labeled data, self-supervised methods have become a major trend, often leveraging the local rigidity of scenes to generate supervisory signals \cite{li2022_rigidflow, baur2021_slim, mittal2020_just_go_with_the_flow}. Diffusion models like DifFlow3D \cite{liu2024difflow3d} have been introduced to improve robustness in noisy environments by modeling motion uncertainty.

Inspired by these successes, researchers have begun to adapt these principles for 4D mmWave radar in automotive scenarios. These works often focus on building robust self-supervised learning strategies by leveraging radar's unique properties. For example, Ding et al. first utilized the Doppler-derived radial velocity for self-supervision \cite{ding2022_selfsupervised_4d_radar} and later explored cross-modal supervision from other sensors like cameras and LiDAR \cite{ding2023_hidden_gems}. Other works like DMRFlow \cite{zhai2025_dmrflow} and TARS \cite{wu2025_tars} have further refined the network architecture for traffic scenes. However, a fundamental limitation shared by both the LiDAR and automotive radar literature is their inherent assumption of a world composed of rigid or piecewise-rigid objects (e.g., vehicles, cyclists). This assumption breaks down when faced with the highly articulated and non-rigid nature of human motion, rendering these methods suboptimal for fine-grained human motion analysis.
\subsection{Motion Sensing and Estimation via Vision and Radar}
The challenge of modeling non-rigid motion has been extensively studied in the computer vision community. Methods using RGB images \cite{bloesch2018codeslam, yang2020upgrading, hur2021_selfsupervised_monocular_scene_flow, zhao2022dit, li2022monoplflownet} have demonstrated the ability to estimate ego motions, dense 3D environments and motion fields. Furthermore, multi-modal approaches fusing RGB with depth or event data have shown increased robustness. CamLiFlow \cite{liu2022camliflow}, for instance, leverages LiDAR's geometric prior to correct image depth errors, while RPEFlow \cite{wan2023rpeflow} and BlinkVision \cite{li2024_blinkvision} integrate event-camera data to better capture high-speed motion. These vision-based methods are effective in capturing deformations of the human body.

However, their reliance on cameras introduces two critical drawbacks for human-centric applications: severe privacy risks and susceptibility to environmental conditions. The intrusive nature of cameras limits their deployment in homes, hospitals, and elderly care facilities. In addition, their performance degrades significantly in poor lighting, smoke, or fog. This fundamental conflict between performance and privacy motivates a shift towards sensing modalities that are both effective and non-invasive. The mmWave radar, with its ability to ``see'' through darkness and preserve anonymity by capturing sparse points instead of detailed appearances, emerges as an ideal solution to bridge this gap.
\subsection{Radar-based Human Motion Sensing and Doppler Prior}
In fact, a large body of research in human sensing has begun to exploit the multi-modal characteristics of mmWave radar to achieve finer-grained perception. These works are no longer limited to processing the final sparse point clouds but delve into the rich features available at different stages of the radar signal processing pipeline. For example, in the cutting-edge area of 3D human mesh reconstruction, works like mmMesh \cite{xue2021_mmmesh} and M4esh \cite{xue2022_m4esh} have demonstrated the remarkable ability to reconstruct dynamic, high-fidelity 3D human meshes directly from radar signals, proving that the raw data contains rich information sufficient to recover detailed body posture and shape. In Human Activity Recognition (HAR) \cite{gu2024millimeter,cao2023_har_feature_attention}, some advanced methods also go beyond point cloud geometry, directly utilizing Range-Doppler Heatmaps \cite{cao2023_har_feature_attention}. Using micro-Doppler effects within this sparse data \cite{yang2024_generalizable_indoor_har,zeng2020automatic}, these models can capture fine-grained motion signatures specific to actions like walking or waving, enabling more robust recognition. These works fully demonstrate that the rich features of the mmWave radar are crucial for a deep understanding of the human body, especially the Doppler information, which greatly inspires us to design a Doppler-informed generation network of human motion flow. Our experimental findings further validate the advantageous role of Doppler information from 4D radars/LiDARs in estimating human motion flow.

    \begin{figure*}[!t]
        \centering
        \includegraphics[width=0.81\textwidth]{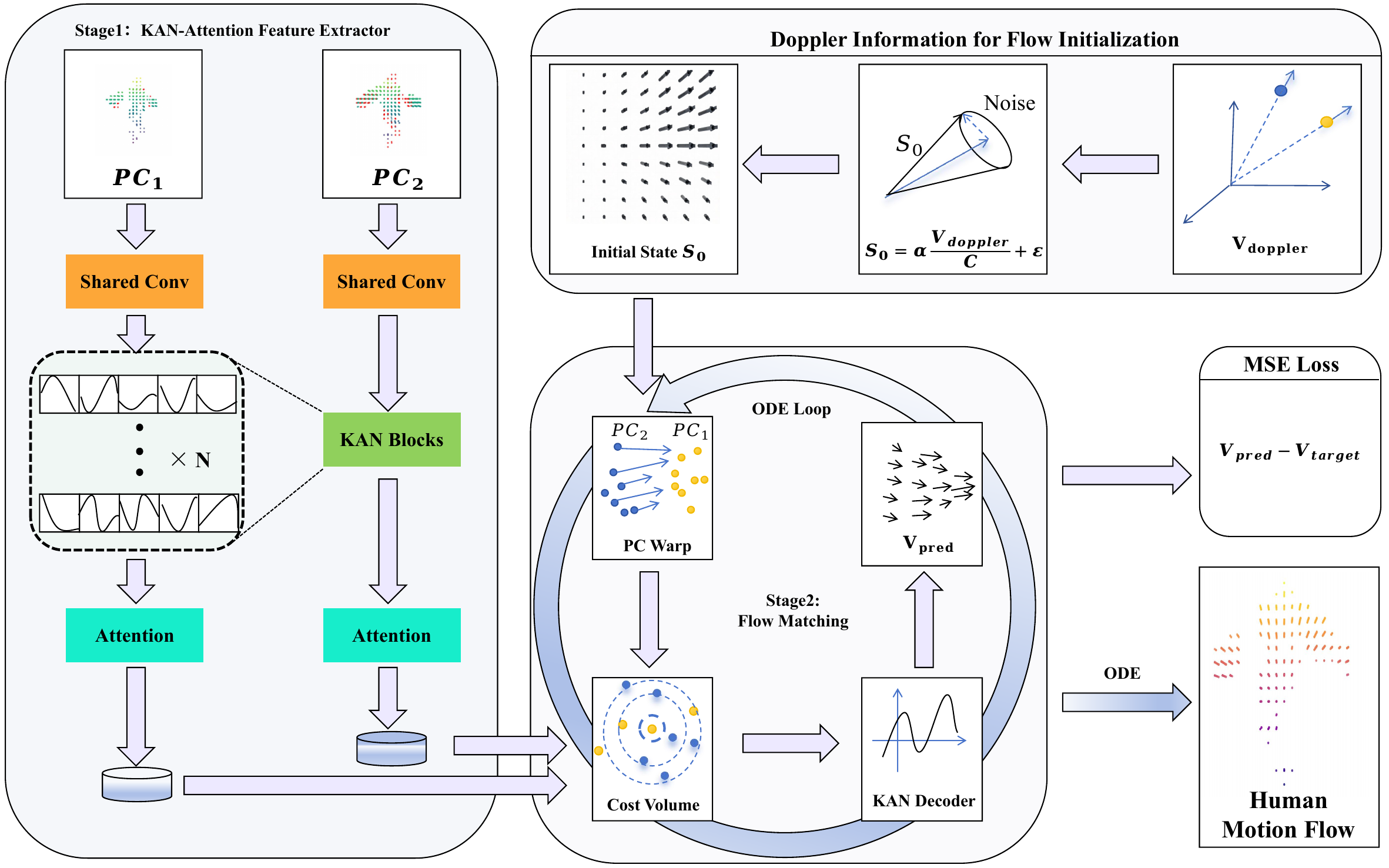}
        \caption{Overview of DiFF. A weight-shared point-wise projection, three KAN blocks, and frame-wise global attention encode the source $\notation{\mathrm{PC}_1}$ and target $\notation{\mathrm{PC}_2}$. During training, the intermediate flow $\mathcal{S}_t$ warps the source points toward the target to construct the state-dependent condition $c_t$. The KAN decoder combines $c_t$, $\mathcal{S}_t$, and the time embedding to predict $v_{\mathrm{pred}}$. During inference, the condition computed at $t=0$ is cached while the ODE updates $\mathcal{S}_t$.}
        \label{fig:architecture}
    \end{figure*}

\section{Methodology} \label{sec:methodology}
\subsection{Problem Formulation}
\revthree{Let $\notation{\mathrm{PC}_1}=\{\mathbf{p}_i\in\mathbb{R}^3\}_{i=1}^{N}$ and $\notation{\mathrm{PC}_2}=\{\mathbf{q}_j\in\mathbb{R}^3\}_{j=1}^{M}$ denote consecutive source and target radar point clouds. The forward ground-truth flow is $\mathcal{S}_{\mathrm{gt}}=\{\mathbf{s}_i\in\mathbb{R}^3\}_{i=1}^{N}$, where $\mathbf{p}_i+\mathbf{s}_i$ is the location of source point $i$ at the target time. This definition does not assume that an observed target return $\mathbf{q}_j$ exists at exactly that location, because radar detections are not persistent across frames. Each source return may also carry intensity, amplitude, and a signed Doppler measurement $d_i$. The task is to estimate $\widehat{\mathcal{S}}=\{\widehat{\mathbf{s}}_i\}_{i=1}^{N}$ on the source points from the two point sets and their available attributes.}
\subsection{Framework Overview}
\revthree{DiFF contains a KAN-attention geometric encoder and a conditional flow-matching backbone (Fig.~\ref{fig:architecture}). The key idea is to use Doppler to initialize the radial part of motion, and then use a state-dependent conditioning signal to guide the recovery of the unobserved tangential components. A training sample is processed as follows:}
    \begin{enumerate}
        \item \revthree{\textit{Geometric Encoding}: A weight-shared encoder maps $\notation{\mathrm{PC}_1}$ and $\notation{\mathrm{PC}_2}$ to point-wise features. It preserves input cardinality, refines local neighborhoods with three KAN blocks, and adds frame context via self attention.}
        
        \item \revthree{\textit{Conditional Flow Matching}: At a sampled path time $t$, the intermediate flow $\mathcal{S}_t$ warps the source points. Cross-frame neighbor aggregation around the warped source returns forms the conditioning signal $c_t$. A dynamic condition uses the current state $\mathcal{S}_t$, while a static condition uses the initial state $\mathcal{S}_0$; this distinction is evaluated in the ablation study. A two-layer KAN decoder then predicts the velocity field $v_{\mathrm{pred}}(\mathcal{S}_t,t,c_t)$.}
    \end{enumerate}
\revthree{The training target is the derivative of a straight path from a Doppler-informed initialization $\mathcal{S}_0$ to $\mathcal{S}_{\mathrm{gt}}$. At inference, the model integrates the learned field from $t=0$ to $t=1$.}
\subsection{Network Architecture}
This section details the architecture of two core modules that are designed to effectively process sparse point clouds and model their spatiotemporal relationships. Compared with rigid-scene point clouds, human radar returns are sparse in local neighborhoods and strongly coupled across distant articulated body parts. This motivates our use of KAN-based local refinement together with global attention.

\subsubsection{KAN-Attention Feature Extractor}
To generate an informative condition vector $y$, we design a single-scale extractor inspired by PointKAN-Elite \cite{shi2025kan}. Unlike traditional hierarchical methods \cite{qi2017_pointnet, qi2017_pointnetpp}, our network maintains full point cloud resolution to prevent critical information loss. It comprises three stages:

\paragraph{Siamese Deep Local Feature Extraction}
The network first employs a shared-weight convolutional feature extractor to process the two input point clouds, generating initial coarse spatial feature representations. These features are then passed through a series of KAN Blocks to extract fine-grained geometric features, as illustrated in Fig.~\ref{fig:architecture}. The design of this module resembles the Local Point Feature (LPF) module in PointKAN \cite{shi2025kan}, but we omit the convolutional steps and directly apply KANs to refine geometric features. During the embedding process, for each neighboring point, we concatenate its relative positional coordinates and features with the features of the source point, expanding the dimension along the neighborhood axis within the point cloud dimension. Point-wise KAN-based refinement is subsequently performed. Each KAN layer utilizes a lightweight Kolmogorov-Arnold network with learnable rational activations, defined as:
    \begin{equation}
    \phi(x) = \text{SiLU}(x) + w \frac{\sum_{i=0}^{m} a_i x^i}{\sqrt{1 + (\sum_{j=1}^{n} b_j x^j)^2}}
    \end{equation}
where $\{a_i\}$ and $\{b_j\}$ are learnable coefficients and $w$ is a learnable scaling factor. Horner's method for polynomial evaluation and explicit gradient calculation accelerates both training and inference. Unlike an MLP with fixed node activations, KAN learns nonlinear functions on edges \cite{liu2024kan}; this provides adaptive local bases for sparse radar neighborhoods, where fine-scale geometry, amplitude changes, and Doppler cues can be smoothed by a standard MLP. This design is consistent with the MLP ablation in Table~\ref{tab:ablation1}.

\paragraph{Global Context Attention}
Following local extraction, features pass through a \textit{Global Attention} module. This standard multi-head self-attention mechanism aggregates information across the entire frame, capturing global structural dependencies essential for recognizing articulated human motion. It allows each point to integrate features from all other points in its cloud, effectively modeling long-range dependencies. This is particularly important for human radar sensing, where sparse limb returns can be locally ambiguous: similar local clusters may correspond to different body parts or motion directions unless interpreted with torso-level and whole-body context.

\subsubsection{Flow Matching Backbone}
The backbone network is responsible for regressing the velocity field. Its processing pipeline consists of the following steps:

\paragraph{PC Warp Module}
In this module, we warp the target point cloud toward the source point cloud to facilitate estimation. For each point in the source point cloud, we identify its neighbors in the warped target point cloud. The relative positions between them are computed and concatenated with their corresponding spatial features and the features of the source point. A key advantage of this design is its dynamic re-evaluation: as the flow field $\mathcal{S}_t$ evolves during the flow matching process, the warping distance changes, allowing the next module to adaptively refine the cost volume. This contributes to stable loss convergence, mitigating the training oscillations often observed with KANs.

\paragraph{Cost Volume Module}
This module closely resembles the KAN Block. After the PC warp module, the aggregated information is processed by another KAN-based aggregator, followed by a max-pooling operation over the neighbor features. While this operation is identical to that in the KAN Block, we omit the final normalization and ReLU activation. We posit that these steps would potentially discard conditional information, which contrasts with the KAN Block's objective of refining geometric features. This process yields a point-wise cost volume that quantifies cross-frame similarity. It is subsequently concatenated with the source point cloud's own features to form the final condition vector $c$. In our notation, the state-dependent version is denoted as $c_t$: dynamic conditioning recomputes this signal from the current flow state $\mathcal{S}_t$, whereas static conditioning fixes it at the initial state $\mathcal{S}_0$.

\paragraph{KAN Decoder Module}
This module fuses the multi-source feature representations to derive the final velocity field. To integrate temporal information, we employ a sinusoidal positional encoding scheme. Both this temporal embedding and the current flow representation $\mathcal{S}_t$ are independently processed by an MLP-based projection layer to achieve dimensional alignment with the conditional vector $c$. The holistic state representation is then constructed via an element-wise summation of the flow embedding, the temporal encoding, and the conditional context. Finally, a two-layer lightweight Kolmogorov-Arnold Network (KAN) serves as the decoder to regress the refined 3D velocity for each point within the current flow state $\mathcal{S}_t$. During inference, we compute the condition once at $t=0$ and cache $c_0$ to reduce repeated neighbor search, while the decoder still receives the evolving $\mathcal{S}_t$ and $t$.
\subsection{Doppler-informed Flow Matching}

A central contribution of our work is the integration of Doppler velocity priors into the Flow Matching framework. Instead of initiating the ODE trajectory from a standard Gaussian noise distribution $\mathcal{N}(0, \mathbf{I})$, we start from a distribution that is already informed by the observed motion, as shown in Fig.~\ref{fig:physical:physical_vision}. Specifically, we define the starting point of our flow trajectory $\mathcal{S}_0$ as:
    \begin{equation}
    \mathcal{S}_0 = \alpha \cdot \frac{\mathbf{V}_{\text{doppler}}}{C} + \epsilon, \quad \text{where} \quad \epsilon \sim \mathcal{N}(0, \sigma^2 \mathbf{I})
    \label{eq:doppler_init}
    \end{equation}
Here, $\mathbf{V}_{\text{doppler}}$ is the Doppler velocity, $C$ is a constant for the radar frame rate (e.g., 30 for Arbe \cite{arbe_robotics_2022}), $\alpha$ is a constant to scaling the mean of the Doppler radial flow (typically set to 1), and $\epsilon$ is a small Gaussian noise to maintain stochasticity.

We adopt the Rectified Flow formulation \cite{liu2022flow}, which defines a straight-line path between the starting flow $\mathcal{S}_0$ and the target ground truth flow $\mathcal{S}_{gt}$:
    \begin{equation}
    \mathcal{S}_t = (1-t)\mathcal{S}_0 + t \mathcal{S}_{gt}
    \label{eq:straight_line}
    \end{equation}
As in \cite{lipman2022flow}, the target velocity for a path with mean $\mu_t$ and standard deviation $\sigma_t$ is given by $v_{target}(\mathcal{S}_t, t) = \frac{\sigma'_t}{\sigma_t}(\mathcal{S}_t - \mu_t) + \mu'_t$. For the straight-line path in \eqref{eq:straight_line}, these terms are:
    \begin{equation}
    \mu_t = (1-t)\frac{\alpha \mathbf{V}_{\text{doppler}}}{C} + t\mathcal{S}_{gt}
    \end{equation}
    \begin{equation}
    \sigma_t = (1-t)\sigma_0 + t\sigma_1 = (1-t)\sigma 
    \end{equation}
    \begin{equation}
    \mu'_t = \frac{d\mu_t}{dt} = \mathcal{S}_{gt} - \frac{\alpha \mathbf{V}_{\text{doppler}}}{C}
    \end{equation}
    \begin{equation}
    \sigma'_t = \frac{d\sigma_t}{dt} = -\sigma
    \end{equation}
Substituting these into the general formula, the target velocity field that our network $v_{pred}$ is trained to regress simplifies to the time derivative of the path:
    \begin{equation}
    v_{target} = \frac{d\mathcal{S}_t}{dt} = \mathcal{S}_{gt} - \mathcal{S}_0
    \end{equation}
The training objective is to minimize the Mean Squared Error between the network's prediction and this target velocity:
    \begin{equation}
    \mathcal{L} = \mathbb{E}_{t, \mathrm{PC}_1, \mathrm{PC}_2} \left[ \| v_{pred}(\mathcal{S}_t, t, c_t) - (\mathcal{S}_{gt} - \mathcal{S}_0) \|^2 \right]
    \end{equation}
By initializing the flow trajectory closer to the final solution, this Doppler-informed approach provides a strong inductive bias, significantly accelerating the training process and improving the final accuracy of the scene flow estimation. At inference, the cached condition $c_0$ is used for efficiency, while $\mathcal{S}_t$ and $t$ still evolve through the ODE solver.


\section{Experiments} \label{sec:experiment}

\subsection{Experimental Setup}

\subsubsection{Implementation Details}
\paragraph{Datasets}
All experiments are conducted on two datasets: the milliFlow dataset \cite{ding2024_milliflow} and the mmBody dataset \cite{chen2022mmbody}. The milliFlow dataset provides sparser and noisier point clouds with only a single feature (``intensity''). The mmBody dataset is a large-scale, high-quality multi-modal dataset designed for 3D human mesh reconstruction from millimeter-wave radar. The employed Arbe radar \cite{arbe_robotics_2022} is equipped with a 48$\times$48 antenna array, which surpasses many commercial single-chip radars and provides rich point clouds where each point carries features including intensity, Doppler velocity, and amplitude. To focus on the core challenge of non-rigid motion estimation, we use only the non-occluded sequences. We adopt a 4:1 train-test split, using sequences 0-5, 7-16 for training and sequences 6, 17-19 for testing, ensuring that subjects in the test set are unseen during training. We use a subset to evaluate model generalization.

\paragraph{Data Processing and Ground Truth Generation}
The mmBody dataset provides not only rich radar point clouds, often exceeding 6000 points per frame, but also synchronized 2D images, SMPL-X 3D meshes, and skeletal poses. As the raw data contains significant environmental clutter, we first filter the point cloud for each frame by retaining only points within a 0.15~m radius of the 3D human SMPL-X skeleton. Each of these points is then assigned a semantic body part label based on the nearest skeleton. Following the automatic annotation technique from milliFlow \cite{ding2024_milliflow}, we compute the SE(3) transformation matrix for each body part between two consecutive frames. The ground truth flow $\mathcal{S}_{gt}$ for each point $p_i$ on a body part $j$ is then derived using the kinematic transformation $s_i = (T_j \circ p_i) - p_i$. For training, we process pairs of frames and uniformly sample or pad the filtered point clouds to a fixed size of $N=512$. During inference, we do not perform sampling and directly estimate scene flow on the variable-sized point clouds.

    \begin{figure}[t!]
        \centering
        \includegraphics[width=0.47\textwidth]{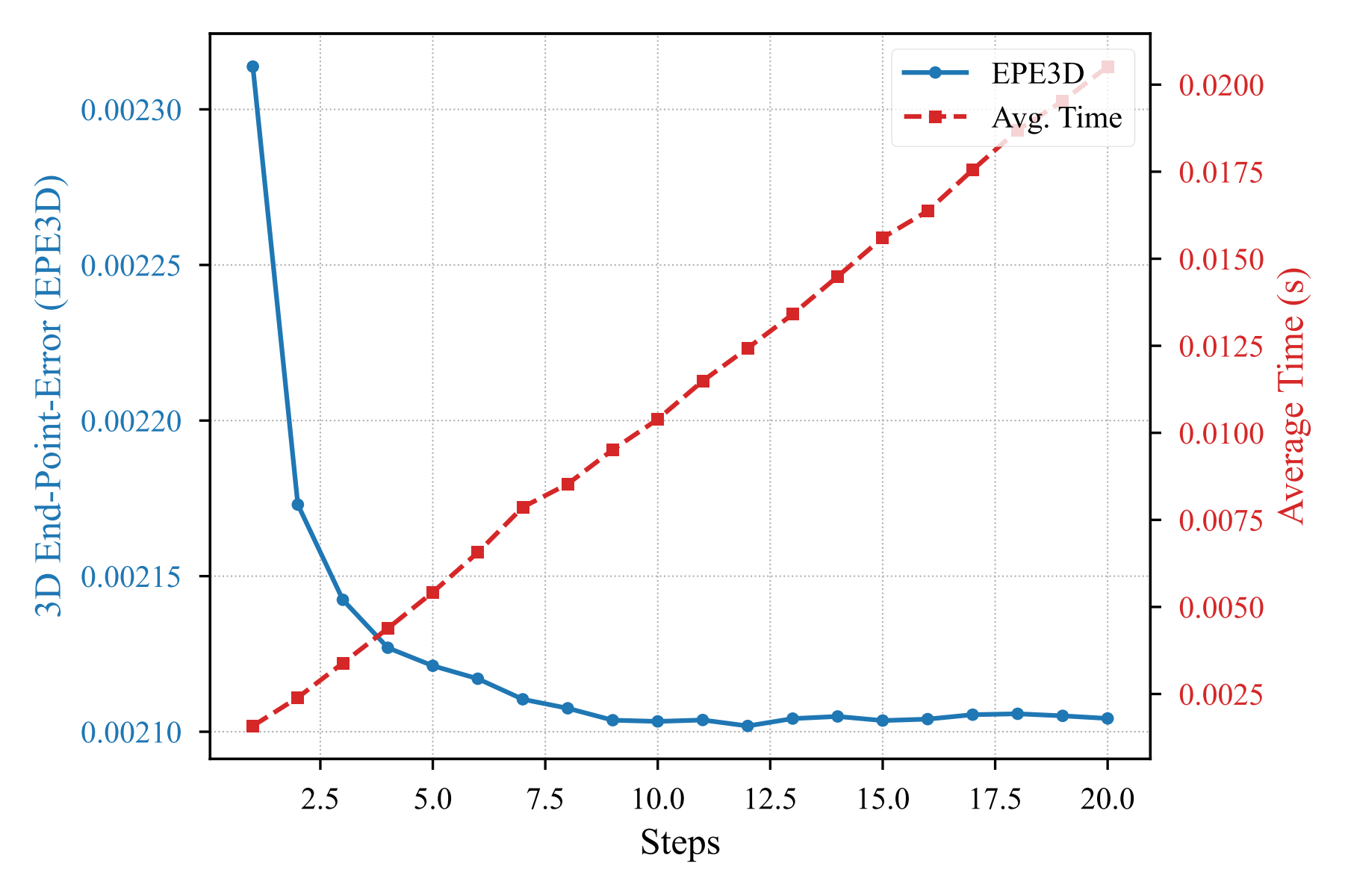}
        \caption{The effect of inference steps on EPE3D and average inference time}
        \label{fig:inference_steps_effect}
    \end{figure}

{\revcolor{dayfivecolor}
    \begin{table*}[t]
        \centering
        \caption{Quantitative comparison on the milliFlow dataset. The cumulative error distribution is shown in Fig.~\ref{fig:combined_epe_distributions} (Left). Since milliFlow does not provide Doppler measurements, the DiFF row follows the radial-prior construction described in the experimental setup. Bold marks the best aggregate value, including ties.}
        \label{tab:milliflow_performance}
        \scriptsize
        \setlength{\tabcolsep}{3.2pt}
        \begin{tabular}{lccccccccc}
        \toprule
        \textbf{Model} & \shortstack{\textbf{EPE3D}\\\textbf{(mm)} $\downarrow$} & \shortstack{\textbf{Acc3D}\\\textbf{Strict (\%)} $\uparrow$} & \shortstack{\textbf{Acc3D}\\\textbf{Relax (\%)} $\uparrow$} & \shortstack{\textbf{0--1 mm}\\\textbf{(\%)}} & \shortstack{\textbf{1--10 mm}\\\textbf{(\%)}} & \shortstack{\textbf{10--100 mm}\\\textbf{(\%)}} & \shortstack{\textbf{100--1000 mm}\\\textbf{(\%)}} & \shortstack{\textbf{$>$1000 mm}\\\textbf{(\%)}} & \shortstack{\textbf{Time}\\\textbf{(ms)} $\downarrow$} \\
        \midrule
        FlowNet3D \cite{liu2019_flownet3d} & 1198.131 & 0.00 & 0.00 & 0.00 & 0.00 & 0.00 & 35.56 & 64.44 & 11.43 \\
        DiffSF \cite{zhang2024_diffsF} & 787.278 & 1.22 & 1.38 & 0.01 & 0.01 & 0.46 & 74.33 & 25.19 & 32.87 \\
        FLOT \cite{puy2020_flot} & 200.508 & 1.54 & 8.53 & 0.00 & 0.00 & 22.53 & 77.38 & 0.09 & 11.07 \\
        milliFlow \cite{ding2024_milliflow} & 177.132 & 0.00 & 0.01 & 0.00 & 0.00 & 0.00 & 99.99 & 0.01 & 30.72 \\
        DifFlow3D \cite{liu2024difflow3d} & 120.295 & 74.39 & 86.51 & 0.01 & 31.24 & 60.92 & 4.71 & 3.12 & 84.63 \\
        FlowStep3D \cite{kittenplon2021_flowstep3d} & 84.264 & 29.63 & 55.24 & 0.00 & 2.66 & 75.01 & 22.32 & 0.01 & 17.83 \\
        DiFF (Ours) & \textbf{27.119} & \textbf{74.61} & \textbf{86.81} & 4.07 & 29.71 & 62.75 & 3.46 & 0.01 & \textbf{19.26} \\
        \bottomrule
        \end{tabular}
    \end{table*}
}

{\revcolor{dayfivecolor}
    \begin{table*}[t]
        \centering
        \caption{Quantitative comparison on the mmBody dataset. The cumulative error distribution is shown in Fig.~\ref{fig:combined_epe_distributions} (Right). Lower EPE3D and inference time are better; higher Acc3D values are better. Bold marks the best aggregate value, including ties.}
        \label{tab:mmbody_comparison_detailed}
        \scriptsize
        \setlength{\tabcolsep}{3.2pt}
        \begin{tabular}{lccccccccc}
        \toprule
        \textbf{Model} & \shortstack{\textbf{EPE3D}\\\textbf{(mm)} $\downarrow$} & \shortstack{\textbf{Acc3D}\\\textbf{Strict (\%)} $\uparrow$} & \shortstack{\textbf{Acc3D}\\\textbf{Relax (\%)} $\uparrow$} & \shortstack{\textbf{0--1 mm}\\\textbf{(\%)}} & \shortstack{\textbf{1--10 mm}\\\textbf{(\%)}} & \shortstack{\textbf{10--100 mm}\\\textbf{(\%)}} & \shortstack{\textbf{100--1000 mm}\\\textbf{(\%)}} & \shortstack{\textbf{$>$1000 mm}\\\textbf{(\%)}} & \shortstack{\textbf{Time}\\\textbf{(ms)} $\downarrow$} \\
        \midrule
        DiffSF \cite{zhang2024_diffsF} & 319.4 & 4.42 & 4.60 & 0.06 & 0.39 & 7.32 & 92.19 & 0.04 & 29.90 \\
        milliFlow \cite{ding2024_milliflow} & 80.2 & 0.03 & 0.20 & 0.00 & 0.00 & 76.81 & 23.19 & 0.00 & 33.30 \\
        FLOT \cite{puy2020_flot} & 30.4 & 47.44 & 86.26 & 0.00 & 4.03 & 95.47 & 0.50 & 0.00 & 19.05 \\
        FlowNet3D \cite{liu2019_flownet3d} & 14.2 & 86.01 & 90.34 & 0.96 & 41.83 & 57.21 & 0.00 & 0.00 & 35.50 \\
        FlowStep3D \cite{kittenplon2021_flowstep3d} & 13.7 & 88.96 & 95.49 & 0.00 & 63.50 & 36.09 & 0.42 & 0.00 & 22.83 \\
        DifFlow3D \cite{liu2024difflow3d} & 2.8 & \textbf{98.63} & 99.62 & 2.67 & 92.24 & 5.09 & 0.00 & 0.00 & 89.82 \\
        DiFF (Ours) & \textbf{2.1} & \textbf{98.63} & \textbf{99.63} & 63.36 & 31.78 & 4.86 & 0.00 & 0.00 & \textbf{10.10} \\
        \bottomrule
        \end{tabular}
    \end{table*}
}

    \begin{figure*}[t!]
        \centering
        \includegraphics[width=0.81\textwidth]{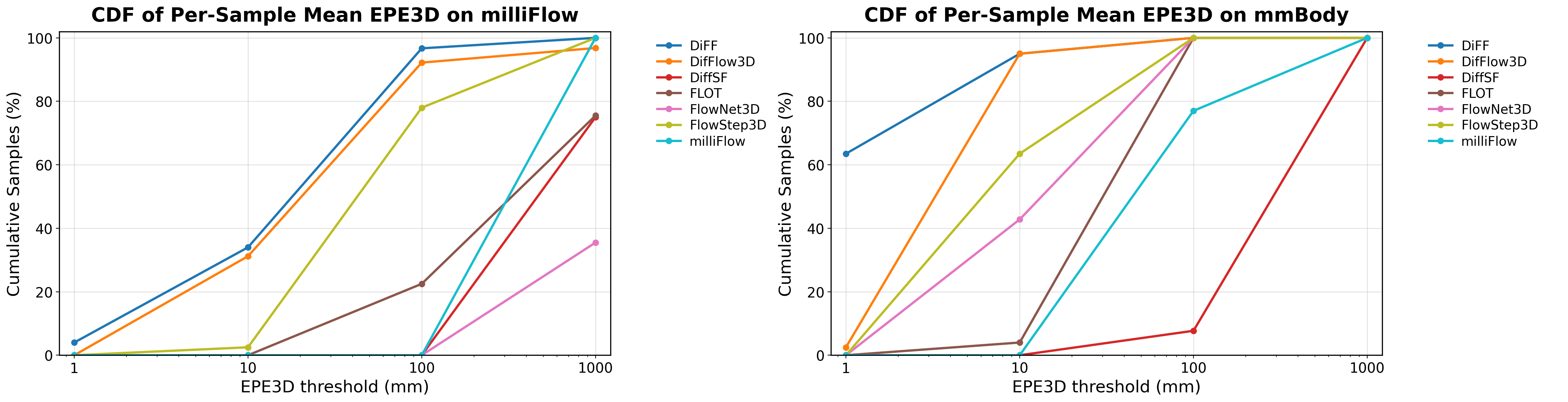}
        \caption{Cumulative distributions of per-sample EPE3D on (Left) milliFlow~\cite{ding2024_milliflow} and (Right) mmBody~\cite{chen2022mmbody}, corresponding to Tables~\ref{tab:milliflow_performance} and \ref{tab:mmbody_comparison_detailed}, respectively.}
        \label{fig:combined_epe_distributions}
    \end{figure*}

\subsubsection{Training and Inference Details}

\paragraph{Doppler Prior Generation}

\revfour{For milliFlow, which does not contain Doppler measurements, we follow the first-submission diagnostic setting and construct radial priors by projecting available flow annotations onto radial directions. This setting allows us to analyze the behavior of the flow-matching backbone on sparser radar point clouds; a fully deployable no-Doppler setting can be addressed by replacing this prior with a learned or uninformed initialization.}

\revfour{For mmBody training, the radial component is obtained by projecting the labeled flow onto the radar line of sight: $\mathbf{s}_{\mathrm{rad},i}=(\mathbf{s}_i^{\top}\widehat{\mathbf{r}}_i)\widehat{\mathbf{r}}_i$, and we set $\mathbf{s}_i^{D}=\mathbf{s}_{\mathrm{rad},i}$ in \eqref{eq:doppler_init}. The random component is sampled with a scaler parameter, and the initial trajectory preserves the physically observed radial motion while still allowing exploration of the unobserved tangential components. At mmBody inference, $\mathbf{s}_i^{D}$ is computed from the measured Doppler velocity and frame interval, with $\boldsymbol{\epsilon}_{\perp,i}=\mathbf{0}$.}

\paragraph{Conditioning}
\revfour{The conditioning signal refers to the cross-frame cost-volume feature used by the velocity decoder. In the static variant, this signal is computed from the initial state $\mathcal{S}_0$, i.e., the Doppler-informed flow at $t=0$. In the dynamic variant used for training, the source points are warped by the sampled intermediate state $\mathcal{S}_t$, and the condition $c_t$ is recomputed from the corresponding target neighbors. Thus static conditioning uses the input at $t=0$, whereas dynamic conditioning uses the time-dependent state along the flow path. During inference, we compute $c_0$ once and cache it to avoid repeated neighbor searches, while the decoder still receives the evolving $\mathcal{S}_t$ and $t$.}

\paragraph{Parameter Settings}
\revfour{The PyTorch model uses 64-dimensional encoder features, three KAN blocks, a 128-dimensional condition and state embedding, and a two-layer KAN decoder. We train end-to-end on one NVIDIA RTX 4080 GPU for 200 epochs with batch size 16, AdamW, an initial learning rate of $1 \times 10^{-3}$, and cosine annealing. Inference uses 10 ODE steps.}
\subsection{Comparison with State-of-the-Art Methods}

We compare DiFF with several SOTA methods, including the leading radar-based approach for human motion, and several strong LiDAR-based methods retrained on the milliFlow and mmBody datasets, as reported in Tables~\ref{tab:milliflow_performance} and~\ref{tab:mmbody_comparison_detailed}.

\paragraph{Evaluation Metrics}
For a comprehensive evaluation, we employ four key metrics: 3D End-Point-Error (EPE3D) in mm, Strict 3D Accuracy (Acc3D Strict) in (\%), Relaxed 3D Accuracy (Acc3D Relax) in (\%), and the EPE3D distribution across test samples. To rigorously assess the model's performance on fine-grained human motion, we adopt significantly stricter accuracy thresholds than those established in the autonomous driving domain. Besides, EPE3D distribution of test samples serves as a key indicator of an algorithm's robustness, reflecting its ability to mitigate the effects of noise inherent in sparse mmWave radar point clouds.These evaluation metrics are defined as follows:
    \begin{itemize}
        \item \textbf{EPE3D (mm)}: The average L2 endpoint error between the predicted and ground-truth scene-flow vectors over all evaluated points. Lower values indicate more accurate scene-flow estimation.
    
        \item \textbf{Acc3D Strict (\%)}: For each test sample, this metric measures the proportion of points whose point-wise endpoint error is no greater than $0.025$ m or whose relative error, normalized by the magnitude of the corresponding ground-truth flow vector, is no greater than $2.5\%$. The reported value is averaged over all test samples. Higher values are better.
    
        \item \textbf{Acc3D Relax (\%)}: This metric is calculated in the same manner as Acc3D Strict, but uses relaxed thresholds of $0.05$ m for the point-wise endpoint error and $5.0\%$ for the relative error. The reported value is averaged over all test samples. Higher values are better.
    
        \item \textbf{EPE3D Distribution}: For each test sample, we compute the mean endpoint error across evaluated points and assign it to one of five ranges: 0–1 mm, 1–10 mm, 10–100 mm, 100–1000 mm, or above 1000 mm. A larger share in lower-error ranges indicates greater accuracy and consistency.
    \end{itemize}
Beyond accuracy, our comprehensive evaluation also considers computational efficiency. As these demanding metrics highlight our algorithm's SOTA performance, our analysis of the average inference time further demonstrates its practicality and superiority for real-world applications.

\paragraph{Results}
In Table~\ref{tab:milliflow_performance}, DiFF maintains SOTA performance even with sparser radar points, with 33.78\% of the flow estimates exhibiting an EPE3D below 10~mm, further confirming its robustness. The cumulative error curves in Fig.~\ref{fig:combined_epe_distributions} show that, regardless of the dataset tested, DiFF exhibits stronger robustness compared to others, with a substantial portion of test data achieving errors under 10~mm.

In Table~\ref{tab:mmbody_comparison_detailed}, DiFF outperforms by a significant margin. It reduces the EPE3D to only 2.1~mm, which represents an order-of-magnitude improvement over the previous leading method, milliFlow. While other diffusion-based models such as DifFlow3D~\cite{liu2024difflow3d} achieve competitive accuracy, their flow errors are mostly concentrated in the range of 1-10 mm, and their inference speed is considerably slower. In contrast, DiFF not only achieves a lower EPE3D, but also locates 63.36\% of the samples within the precise 0-1 mm range, demonstrating a clear advantage for real-time applications.

These advancements can be attributed to two key innovations in our method. First, the substantial performance leap is driven by the effective integration of Doppler priors into the generative flow-matching framework. Second, by replacing the cumbersome downsampling backbones (e.g., PointNet++) with an efficient hybrid architecture that leverages KANs and global attention on raw point clouds, we are able to robustly capture both local and global geometric structures in sparse point clouds while mitigating the impact of noise. This architectural efficiency significantly reduces training and inference time, enabling a compelling combination of high accuracy and real-time performance.
\subsection{Ablation Experiment}
To analyze the sources of our model's performance gains, we conduct a series of ablation studies, with results summarized in Table~\ref{tab:ablation1} and Table~\ref{tab:ablation2}.

\paragraph{Effect of Core Components}
Table~\ref{tab:ablation1} validates our key design choices.

{\revcolor{dayfourcolor}
    \begin{itemize}
        \item \textbf{w/o Doppler Prior:} We replace the Doppler-informed initialization with a standard Gaussian noise prior. The sharp performance drop confirms that our Doppler initialization provides a \textbf{strong inductive bias}.
        \item \textbf{Using MLP:} We replace all KAN-based components with standard MLPs. The performance degradation indicates that KAN's \textbf{adaptive activations} are more effective in capturing complex geometric relationships in sparse point clouds.
        \item \textbf{Static Conditioning:} We compare the dynamic cost volume with a training variant that uses static conditioning. The results show that \textbf{dynamic conditioning} provides more robust guidance throughout the ODE solving process.
    \end{itemize}
}

{\revcolor{dayfivecolor}
    \begin{table}[t]
        \caption{Ablation study on core components. Configuration names match the variants defined in Section~\ref{sec:experiment}.}
        \label{tab:ablation1}
        \centering
        \begin{tabular}{lc}
        \toprule
        \textbf{Configuration} & \textbf{EPE3D (mm)} $\downarrow$ \\
        \midrule
        DiFF (Full Model) & \textbf{2.1} \\
        w/o Doppler Prior & 4.4 \\
        Using MLP & 2.7 \\
        Static Conditioning & 3.3 \\
        \bottomrule
        \end{tabular}
    \end{table}
}

\paragraph{Effect of Initial Noise Variance}
We also study the impact of the variance $\sigma^2$ of the noise in the Doppler prior. In Table~\ref{tab:ablation2}, the performance is optimal at $\sigma^2=0.01^2$. A larger variance introduces more noise, making it difficult to learn the path from the Doppler mean to the target flow. Conversely, a smaller variance reduces the model's ability to correct for errors in the initial prior, as the ODE trajectory becomes too deterministic and can be led astray by inaccuracies in the Doppler measurement and velocity errors made in the ODE solver. Notably, the setting with the best EPE3D does not produce the lowest training loss, indicating that training loss alone is not sufficient for selecting the noise variance.

{\revcolor{dayfivecolor}
    \begin{table}[t]
        \caption{Sensitivity to noise variance. The lowest EPE3D is highlighted; training loss is reported for reference and is not minimized at the best-EPE3D setting.}
        \label{tab:ablation2}
        \centering
        \begin{tabular}{ccc}
        \toprule
        \textbf{Variance ($\sigma^2$)} & \textbf{EPE3D (mm)} $\downarrow$ & \textbf{Avg. Training Loss} \\
        \midrule
        $1.0$ & 3.0 & $3.20 \times 10^{-4}$ \\
        $0.1^2$ & 2.3 & $9.60 \times 10^{-5}$ \\
        $\mathbf{0.01^2}$ & \textbf{2.1} & $3.20 \times 10^{-5}$ \\
        $0.001^2$ & 2.4 & $9.60 \times 10^{-6}$ \\
        $0.0001^2$ & 2.6 & $4.80 \times 10^{-6}$ \\
        $0.00001^2$ & 3.0 & $4.80 \times 10^{-6}$ \\
        \bottomrule
        \end{tabular}
    \end{table}
}


\section{Conclusion} 

In this work, we presented DiFF, a novel framework for human motion scene flow estimation from sparse mmWave radar point clouds. By integrating Doppler velocity priors into a flow matching  model and leveraging a KAN-based feature extractor, our approach achieves state-of-the-art performance on the mmBody dataset. Ablation studies confirm the significance of each component: the Doppler prior, the KAN architecture, and the dynamic correlation mechanism.

Despite its strengths, DiFF has limitations that suggest future directions: (1) replacing the linear flow path with a learned, nonlinear transport to better exploit the anisotropic Doppler prior; (2) extending the framework to multi-person scenarios with occlusion handling. Overall, DiFF represents a meaningful step toward human motion analysis with applications in assistive robotics and human-robot interaction.


\bibliographystyle{IEEEtran}
\IEEEtriggeratref{29}
\bibliography{ref}

\end{document}